\documentclass[11pt,a4paper]{article}
\usepackage[hyperref]{emnlp}
\usepackage{times}
\usepackage{latexsym}

\usepackage{url}
\usepackage{graphicx}
\usepackage{multirow}
\usepackage{amsthm}
\usepackage{graphicx}
\usepackage{enumitem}
\usepackage{scalerel}
\usepackage{amsmath}
\usepackage{amssymb}
\usepackage{bbm}
\usepackage{soul}
\usepackage{amsmath}

\usepackage{standalone}
\usepackage{times}
\usepackage{latexsym}
\usepackage{multirow}
\usepackage{amsthm}
\usepackage{graphicx}
\usepackage{enumitem}
\usepackage{scalerel}
\usepackage{amsmath}
\usepackage{amssymb}
\usepackage{flushend}
\usepackage{tkz-berge}

\aclfinalcopy 

\DeclareMathOperator*{\relu}{relu}
\DeclareMathOperator*{\sigmoid}{sigmoid}
\DeclareMathOperator*{\argmax}{arg\,max}

\title{Combining Spans into Entities:\\ A Neural Two-Stage Approach for Recognizing Discontiguous Entities }

\author{Bailin Wang \\
ILCC, School of Informatics \\ 
University of Edinburgh \\
{\tt bailin.wang@ed.ac.uk} \\\And
Wei Lu \\ 
StatNLP Research Group \\
Singapore University of Technology and Design \\
{\tt luwei@sutd.edu.sg}}

\date{}

\begin{document}
\maketitle
\begin{abstract}

In medical documents, it is possible that an entity of interest not only contains a discontiguous sequence of words but also overlaps with another entity. 
Entities of such structures are intrinsically hard to recognize due to the large space of possible entity combinations. 
In this work, we propose a neural two-stage approach to recognizing discontiguous and overlapping entities by decomposing this problem into two subtasks: 
1) it first detects all the overlapping spans that either form entities on their own or present as segments of discontiguous entities, based on the representation of segmental hypergraph, 
2) next it learns to combine these segments into discontiguous entities with a classifier, which filters out other incorrect combinations of segments.  
Two neural components are designed for these subtasks respectively and they are learned jointly using a shared encoder for text. 
Our model achieves the state-of-the-art performance in a standard dataset, 
even in the absence of external features that previous methods used.

\end{abstract}

\section{Introduction}

Named entity recognition (NER) aims at identifying shallow semantic elements in text and has been a crucial step towards natural language understanding \cite{tjong2003introduction}. 
Extracted entities can facilitate various downstream tasks like  question answering \cite{abney2000answer}, relation extraction \cite{mintz2009distant,liu2017heterogeneous}, event extraction \cite{riedel2011fast,lu2012automatic,li-ji-huang:2013:ACL2013}, and coreference resolution \cite{soon2001machine,ng2002improving,chang2013constrained}.

The underlying assumptions behind most NER systems are that an entity should contain a contiguous sequence of words and should not overlap with each other. 
However, such assumptions do not always hold in practice.
First, entities or mentions\footnote{Mentions are defined as references to entities that could be named, nominal or pronominal \cite{florian2004statistical}.}
with overlapping structures frequently exist in news \cite{doddington2004automatic} and  biomedical documents \cite{kim2003genia}.
Second, entities can be discontiguous, especially in clinical texts \cite{pradhan2014evaluating}. 
For example,  Figure \ref{fig:example} shows three entities where two of them are discontiguous (``{\em laceration \dots esophagus}'' and ``{\em stomach \dots lac}''), and the second discontiguous entity also overlaps with another entity (``{\em blood in stomach}'').


\begin{figure}[t!]


\includegraphics[width=7.7cm]{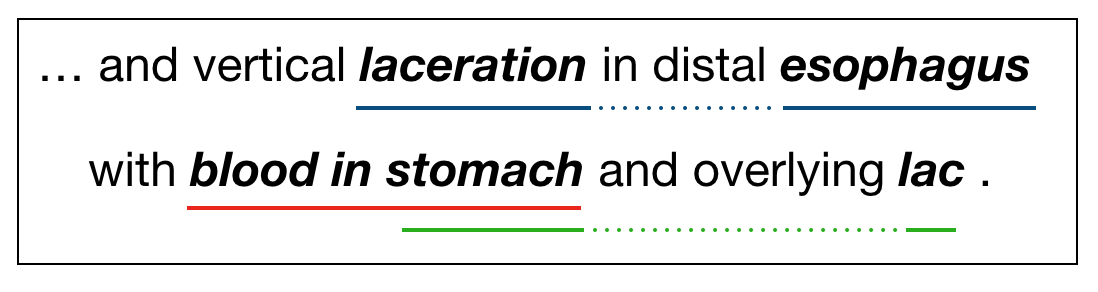}
\caption{Entities are highlighted with colored underlines. ``{\em laceration ... esophagus}" and ``{\em stomach ... lac}" contain discontiguous sequence of words and the latter also overlaps with another entity ``{\em blood in stomach}". }
\label{fig:example}
\end{figure}

Such discontiguous entities are intrinsically hard to recognize considering the large search space of possible combinations of entities that have discontiguous and overlapping structures.
\citet{muis2016learning} proposed a hypergraph-based representation to compactly encode discontiguous entities.
However, this representation suffers from the ambiguity issue during decoding -- one particular hypergraph corresponds to multiple interpretations of entity combinations.
As a result, it resorted to heuristics to deal with such an issue. 

Motivated by their work, we take a novel approach to resolve the ambiguity issue in this work.
Our core observation is that though it is hard to exactly encode the exponential space of all possible discontiguous entities,
recent work on extracting overlapping structures \cite{D18-1019} can be employed to efficiently explore the space of all the span combinations of discontiguous entities. 
Based on this observation, we decompose the problem of recognizing discontiguous entities into two subtasks: 
1) {\em segment extraction}: learning to detect all (potentially overlapping) spans that either form entities on their own or present as parts of a discontiguous entity; 
2) {\em segment merging}: learning to form entities by merging certain spans into discontiguous entities.

Our contributions are summarized as follows:

\begin{itemize}[noitemsep,topsep=0pt,parsep=0pt,partopsep=0pt]
\vspace{2mm}
\item By decomposing the problem of extracting discontiguous entities into two subtasks, 
we propose a two-stage approach that does not have the ambiguity issue. 

\vspace{2mm}
\item Under this decomposition, we design two neural components for these two subtasks respectively.
We further show that the joint learning setting where the two components use a shared text encoder is beneficial. 

\vspace{2mm}
\item Empirical results show that our system achieves a significant improvement compared with previous methods, 
even in the absence of external features that previous methods used.
\footnote{Our code is available at {\url{https://github.com/berlino/disco_em19}}.}

\end{itemize}
\vspace{2mm}

Though we only focus on discontiguous entity recognition in this work, our model may find applications in other tasks that involve discontiguous structures, such as detecting gappy multiword expressions \cite{schneider2014discriminative}.

\section{Related Work}

The task of extracting overlapping entities has long been studied \cite{zhang2004enhancing,zhou2004recognizing,zhou2006recognizing,mcdonald2005flexible,alex2007recognising,finkel2009nested,lu2015joint,muis2017labeling}.
As neural models \cite{collobert2011natural,lample2016neural,huang2015bidirectional,chiu2016named,ma-hovy:2016:P16-1} are proven effective for NER, 
there have been several neural systems
recently proposed to handle entities of overlapping structures
 \cite{N18-1131,N18-1079,D18-1124,D18-1309,D18-1019,strakova-etal-2019-neural,lin-etal-2019-sequence,fisher-vlachos-2019-merge}. 
Our system is based on the model of neural segmental hypergraphs \cite{D18-1019} 
which encodes all the possible combinations of overlapping entities using a compact hypergraph representation without ambiguity.
Note that other system for extracting overlapping structures can also fit into our two-stage system. 
 
For discontiguous and overlapping entity recognition, 
\citet{tang2013recognizing,zhang2014uth_ccb,xu2015uth} extended the BIO tagging scheme to encode such complex structures 
so that traditional linear-chain CRF \cite{lafferty2001conditional} can be employed.
However, the model suffers greatly from ambiguity during decoding due to the use of the extended tagset.
\citet{muis2016learning} proposed a hypergraph-based representation to reduce the level of ambiguity.
Essentially, these systems trade expressiveness for efficiency: 
they inexactly encoded the whole space of discontiguous entities with ambiguity for training, 
and then relied on some heuristics to handle the ambiguity during decoding.
\footnote{We will briefly introduce them and their heuristics later as our baselines in our experiments for comparison.}
Considering it is intrinsically hard to exactly identify discontiguous entities in one stage using a structured model, 
our work tries to decompose the task into two sub-tasks to resolve the ambiguity issue.

This task is also related to joint entity and relation extraction \cite{kate2010joint,li2014incremental,miwa2014modeling} 
where the discontiguous entities can be viewed as relation links between segments.
The major difference is that discontiguous entities require explicitly modeling overlapping entities and linking multiple segments.

\section{Model}

Our goal is to extract a set of entities that may have overlapping and discontiguous structures given a natural language sentence.
We use $ \boldsymbol x = x_1\dots x_{|\boldsymbol{x}|}$ to denote a sentence, and use 
$\boldsymbol y = \{ [b_{i:j} \dots b_{m:n} ]^k \} $ to denote a set of discontiguous entities 
where each entity of type $k$ contains a list of spans, e.g., $b_{i:j}$ and $b_{m:n}$, 
with subscripts indicating the starting and ending positions of the span. 
Hence, this task can be viewed as extracting and labelling a sequence of spans as an entity.

Our two-stage approach first extracts spans of interest like $b_{i:j}$, which are parts of discontiguous entities.
Then it merges these extracted spans into discontiguous entities.
In the more general setting where discontiguous entities are typed,
our approach is designed to jointly extract and label the spans at the first stage, then only merge the spans of the same type at the second stage.
We call the intermediate typed span $b_{i:j}^k $ a \textit{segment} in the rest of the paper. 

Formally, our model aims at maximizing the conditional probability $p(\boldsymbol y| \boldsymbol x)$, which is decomposed as:
\begin{align}
p(\boldsymbol y| \boldsymbol{x} ) = p(\boldsymbol s| \boldsymbol x)p( \boldsymbol y| \boldsymbol s, \boldsymbol x)
\end{align}

\noindent where $ \boldsymbol s= \{ b_{i:j}^k \} $ denotes the set of segments that leads to $\boldsymbol y$ through a specific combination.
\footnote{We note that each $\boldsymbol y$ corresponds to one unique $\boldsymbol s$.}
That is, we divide the problem of extracting discontiguous entities into two subtasks, 
namely \textit{segment extraction} and \textit{segment merging}.

\subsection{Segment Extraction}

The entity segments $\boldsymbol s$ of interest in a given sentence could also overlap with each other.
For example, in Figure \ref{fig:example} the entity ``{\em blood in stomach}" contains another segment ``{\em stomach}".
To make our model capable of extracting such overlapping segment combinations, 
we employ the model of neural segmental hypergraphs from \citet{D18-1019}, which uses a hypergraph-based representation to encode all the possible combinations of segments without ambiguity. 
Specifically, the segmental hypergraphs adopt a log-linear approach to model the conditional probability of each segment combination for a given sentence:

\begin{equation}
p(\boldsymbol{s} |\boldsymbol{x}) = \frac{\exp  f(\boldsymbol{x},\boldsymbol{s})}{\sum_{\boldsymbol{s}'} \exp f(\boldsymbol{x},\boldsymbol{s}')}
\label{eq:loglinear}
\end{equation}
where $f(\boldsymbol{x},\boldsymbol{s})$ is the score function for any pair of input sentence $\boldsymbol{x}$ and output segment combination $\boldsymbol s$.

In segmental hypergraphs, each segment combination $\boldsymbol s$ corresponds to a hyperpath.
Following \citet{D18-1019}, the score for a hyperpath is the sum of the scores for each hyperedge, 
which are based on the word-level and span-level representations through LSTM \cite{graves2005framewise}:
\begin{align}
\boldsymbol{}\mathbf{h}_i^w = [{{\mathop{\rm biLSTM}\nolimits} _1}({\mathbf{x}}_0,...,{\mathbf{x}}_n)]_i
\label{eq:word}
\\
\boldsymbol{}{\bf{h}}_{i:j}^s = {{\mathop{\rm biLSTM}\nolimits} _2}({\mathbf{h}}_i^w,...,{\mathbf{h}}_j^w)
\label{eq:span}
\end{align}
where $\mathbf{x}_k$ is the corresponding word embedding for word $x_k$, ${\bf{h}}_i^w$ denotes the representation for the $i$-{th} word and ${\bf{h}}_{i:j}^s$ denotes the representation for the span from the $i$-th to the $j$-th word.

On top of the segmental hypergraph representation, 
the partition function which is the denominator of Equation \ref{eq:loglinear} can be computed using dynamic programming.
The inference algorithm has a quadratic time complexity in the number of words, 
which can be further reduced to linear time complexity if we introduce the maximal length $c$ of a segment. 
We regard $c$ as a hyperparameter.

\begin{table}
\centering
{\def\arraystretch{1.15}\tabcolsep=2pt
\scalebox{0.8}{
\begin{tabular}{c|c|c|c|c|rr}
 & \multirow{2}{*}{\# \textit{sents}} & \multicolumn{3}{c|}{\# {\em entities}  (\%) } & \multirow{2}{*}{\# \textit{o.l.} (\%) }\\
& & \multicolumn{1}{c|}{1 {\em segment}} & \multicolumn{1}{c|}{2 {\em segments}} & \multicolumn{1}{c|}{3 {\em segments}} &  \\\hline
Train & 534 & 544  (46) & 607 (51) & 44 (4) & 205 (17) \\
Dev & 303 & 357  (45) & 421 (53) & 18 (2)  & 240 (30) \\
Test & 430 & 584  (48) & 610 (50) & 16 (1)  & 327 (27)\\
\end{tabular}
}
}
\caption{Statistics of the dataset. \textit{o.l.}: overlapping entities, \textit{sents}: sentences.}
\label{stats}
\end{table}

\subsection{Segment Merging}

Given a set of segments, our next subtask is to merge them into entities. 
First, we enumerate all the valid segment combinations, denoted as $\mathbf E$,
based on the assumption that the segments in the same entity should have the same type and not overlap with each other.
Our model then independently decides whether each valid segment combination forms an entity.
We call these valid segment combinations \textit{entity candidates}.
For brevity, let us use $\boldsymbol{t} ^k$ to denote an entity candidate $ {[b_{i:j} \dots b_{m:n} ]^k}$ where each segment like $b_{i:j}^k$ belongs to $\boldsymbol s $. 

Formally, given segments $\boldsymbol s$ , the probability of generating entities $\boldsymbol y$ can be represented as:

\begin{multline}
p( \boldsymbol y| \boldsymbol s, \boldsymbol x) = 
\prod_{  { \boldsymbol t ^k} \in \mathbf{E}}  \big ( \mathbbm 1[ { \boldsymbol t ^k}  \in \boldsymbol y ]   p(  { \boldsymbol t ^k} \in \boldsymbol y )  \\ 
+  \mathbbm 1[ { \boldsymbol t ^k}  \notin \boldsymbol y ]   ( 1- p(  { \boldsymbol t ^k} \in \boldsymbol y ) )
\big   )
\label{eq:segs}
\end{multline}
where $\mathbbm 1$ is an indicator function. 
We use a binary classifier to model $p(  { \boldsymbol t ^k} \in \boldsymbol y )$.

To capture the interactions between segments within the same combination, we employ yet another LSTM on top of segments as follows:
\begin{equation}
{\bf{h}}_{ {\boldsymbol{t}^k } }^e = {{\mathop{\rm biLSTM}\nolimits} _3}({\bf{h}}_{i:j}^s,...,{\bf{h}}_{m:n}^s)
\label{eq:seg_rep}
\end{equation}
where ${\bf{h}}_{ {\boldsymbol{t}^k } }^e $ denotes the representation of the segment combination ${\boldsymbol{t}^k }  $, 
which then serves as a feature vector for a binary classifier to determine whether it is an entity.
Note that we reuse the span representation from Equation \ref{eq:span}, meaning that encoder for words and spans are shared in both segment extraction and merging.

The binary classifier for each ${\boldsymbol{t}^k }$ in Equation \ref{eq:segs} is computed as:
\begin{equation}
p(  { \boldsymbol t ^k} \in \boldsymbol y )= \sigmoid( \mathbf{W} \cdot \relu ( {\bf{h}}_{ {\boldsymbol{t}^k }}^e) + b )
\end{equation}
where we use a rectified linear unit (ReLU) \cite{glorot2011deep} and a linear layer, parameterized by $\mathbf{W}$ and $ b$,  to 
map the representation from Equation \ref{eq:seg_rep} to a scalar score.
This score is normalized into a distribution by the $\sigmoid$ function.

In the joint model, 
we stack three separate LSTMs to encode text at different levels
from words to spans, then discontiguous entities.
Intuitively, the word and span level LSTM try to capture the lower-level information for segment extraction 
while the entity level LSTM captures the higher-level information for segment merging.

\subsection{Learning and Decoding}

For a dataset $\mathcal D$ consisting of sentence-entities pairs $(\boldsymbol{x}, \boldsymbol{y})$, 
our objective is to minimize the negative log-likelihood as follows:
\begin{equation}
\mathcal L(\theta) =  - \sum_{(\boldsymbol{x}_i, \boldsymbol{y}_i) \in \mathcal D}  \log p(\boldsymbol{y}_i| \boldsymbol{x}_i )+ \frac{\lambda}{2} \Vert \theta \Vert ^2 
\end{equation}
where $\theta$ denotes model parameters and $\lambda$ is the $\ell_2$ coefficient.
$p(\boldsymbol{y}_i | \boldsymbol{x}_i)$ is computed by $p(\boldsymbol{s}_i | \boldsymbol{x}_i) p(\boldsymbol{y}_i| \boldsymbol{s}_i, \boldsymbol{x}_i) $
where $\boldsymbol{s}_i$ is inferred from $\boldsymbol{y}_i$.

During the decoding stage, 
the system first predicts the most probable segments from the neural segmental hypergraph by 
$ \boldsymbol{\hat{s}} = \argmax_{\boldsymbol{{s'}}}  p(\boldsymbol{s'} |\boldsymbol{x}) $.
Then it feeds the prediction to the next stage of merging segments and outputs the discontiguous entities by
$  \boldsymbol{\hat{y}} = \argmax_{\boldsymbol{y'}} p( \boldsymbol y'| \boldsymbol{\hat{s}}, \boldsymbol x)$.

\section{Experiments}

\subsection{Setup}

\paragraph{Data} We evaluated our model on the task of recognizing mentions
in clinical text from ShARe/CLEF eHealth Evaluation Lab (SHEL) 2013 \cite{suominen2013overview} 
and SemEval-2014 \cite{pradhan2014semeval}.
The task is defined to extract mentions of \textit{disorders} from clinical documents according to the Unified Medical Language System (UMLS).
The original dataset only has a small percentage of discontiguous entities, making it not suitable for comparing the effectiveness of different models when handling discontiguous entities.
Following \citet{muis2016learning}, we use a subset of the original data where each sentence contains at least one discontiguous entity. 

We split the dataset according to the setting of SemEval 2014. 
Statistics are shown in Table \ref{stats}. 
In this subset, 53.6\% of entities are discontiguous.
Overlapping entities also frequently appear.
Since an entity has three segments at most, 
we make the constraint that an entity candidate has no more than three segments during segment merging.

Note that all entities hold the same type of \textit{disorder} in this dataset.
Our model is intrinsically able to handle discontiguous entities of multiple types. 
Recall that segments are typed as $b_{i:j}^k$ during segment extraction,
and only segments of the same type can be merged into an entity $ \boldsymbol t ^k$ 
where $k$ indicates the entity type.
To assess its ability to deal with multiple entity types,
we conducted a further analysis (see section \ref{sec:analysis}).  

\paragraph{Hyperparameters} We use the pretrained word embeddings from \citet{chiu2016train} which are trained on the PubMed corpus. 
A dropout layer \cite{srivastava2014dropout} is used after each word is mapped to its embedding. 
The dropout rate and the number of hidden units in LSTMs 
are tuned based on the performance on the development set. 
We set the maximal length of a segment to be 6 during segment extraction.
Our model is trained with Adam \cite{kingma2014adam}.
\footnote{See Appendix C for the full hyperparameters.}

\paragraph{Baselines} 
The first baseline we consider is to extend the traditional BIO tagging scheme to seven tags following \citet{tang2013recognizing}.
With this tagging scheme, each word in a sentence is assigned a label. 
Then a linear-chain CRF is built to model the sequence labelling process.
The next baseline is a hypergraph-based method by \citet{muis2016learning}. 
It encodes each entity combination into a directed graph based on six types of nodes; 
each has its specific semantics.

Since these baselines are both ambiguous,
heuristics are required during decoding. 
Following \citet{muis2016learning}, 
we explored two heuristics: given a model's ambiguous output, either a tag sequence or a hypergraph, 
the ``enough" heuristic finds the minimal set of entities that corresponds to it,
while  ``all"  decodes the union of all the possible set of entities.
Please refer to \citet{muis2016learning} for details. 
We also describe them in the Appendix for self-containedness.

We compare our approach to these baselines in two settings.
In the non-neural setting,
we compare models using the same set of handcrafted features, 
including external features from POS tagger and Brown cluster following \cite{muis2016learning}.
In the neural setting, we implement a linear-chain CRF model using the same neural encoder. 
We are trying to see our model can perform better in both settings.
Note that all neural models in our experiments do not leverage any handcrafted features.

\begin{table}[t]
\centering
\vspace{0mm}
\scalebox{0.9}{
\begin{tabular}{c|l|c|c|c}
& Model &$P$ & $R$ & $F_1$  \\ 
\hline
\multirow{5}{*}{ \shortstack{ \small{Non-} \\ \small{Neural} } } 
& CRF (enough) &  54.7 & 41.2 & 47.0     \\
& CRF (all) &  15.2 & 44.9 &  22.7  \\
& Graph (enough)       &  \textbf{76.9} & 40.1 & 52.7     \\
& Graph (all) &  76.0 & 40.5 & 52.8 \\
& \textit{Our model}  &   76.3   &  41.4 & 53.7  \\
 \hline
 \hline
\multirow{4}{*}{\small{Neural}}
& CRF (enough) & 43.7 &  54.3 & 48.4    \\
& CRF (all)  & 15.7 &  55.8 & 24.5   \\
& \textit{Our model}   & 48.4 & \textbf{66.5} & \textbf{56.1}   \\
& \ \textit{w.o.} \small{\textit{shared encoder}}   & 46.2& 65.1  & 54.0   \\
\end{tabular}
}
\caption{Main results. Graph: the hypergraph based model by \citet{muis2016learning}. ``enough" and ``all" denotes the heuristics used in ambiguous model. 
\textit{w.o. shared encoder}: without using shared encoder.}
\vspace{-2mm}
\label{tab:result}
\end{table}

\subsection{Results and Analysis}\label{sec:analysis}

The main results are listed in Table \ref{tab:result}.
In both non-neural and neural settings, our model achieves the better result in terms of $F_1$ compared with other baselines, 
revealing the effectiveness of our methodology of decomposing the task into two stages.
Our neural model achieves the best performance even without using any external handcrafted features. 

We also assess the performance when our model uses separate encoders for segment extraction and merging. 
From the results, we observe that the setting of using a shared encoder is very beneficial for our two-stage system.

Compared with non-neural models, neural models are better in terms of $F_1$, both for CRF and our models. 
The gain mostly comes from the ability to recall more entities.
Handcrafted features in non-neural models lead to high precisions 
but do not seem to be general enough to recall most entities.

The ``enough" heuristic works better than ``all" in most cases. 
Hence we use it for evaluating models' ability in handling multiple entity types.

\paragraph{Handling Multiple Entity Types} 
To assess the effectiveness of handling entities of multiple types, 
we further categorize each entity into three types based on its Concept Unique Identifier (CUI), 
following \citet{muis2016learning}.
\footnote{Note that the label of CUI is not available for all entities. 
Entities without CUI are assigned with a default \textit{NULL} label.}
In this setting, segments are jointly extracted and labelled using these three categories during segment extraction.
During segment merging, an entity candidate can only contain segments of the same type during merging. 

The results are listed in Table \ref{tab:result2}. 
Our neural model again achieves the best performance among all models in terms of $F_1$.
Compared with neural CRF, our model is significantly better at recalling entities.
Similar to the previous observation, the neural encoder consistently 
boosts the performance of the CRF by recalling more entities, compared with its non-neural counterpart.

\begin{table}[t]
\centering
\vspace{0mm}
\scalebox{0.9}{
\begin{tabular}{c|l|c|c|c}
& Model &$P$ & $R$ & $F_1$  \\ 
\hline
\multirow{2}{*}{ \shortstack{ \small{Non-} \\ \small{Neural} } } 
& CRF (enough) & 55.3 & 37.4 & 44.6     \\
& Graph (enough)    &  \textbf{67.3} & 37.5 & 48.2     \\
 \hline
 \hline
\multirow{2}{*}{\small{Neural}}
& CRF (enough) & 41.6 &  52.3 & 46.3    \\
&  \textit{Our model}  &   43.3   &  \textbf{65.8} & \textbf{52.2}  \\
\end{tabular}
}
\caption{Results on handling multiple entity types.}
\label{tab:result2}
\end{table}

\section{Conclusion and Future Work}

In this work, we propose a neural two-stage approach for recognizing discontiguous entities,
which learns to extract and merge segments jointly without suffering from ambiguity issue.
Empirically, it achieves a significant improvement compared with previous methods 
that rely heavily on handcrafted features.

During training, the classifier of merging segments is only exposed to correct segments, making it unable to recover from errors of segment exaction during decoding.
This issue is similar to \textit{exposure bias} \cite{D16-1137} and
it might be beneficial if the classifier of segment merging is exposed to 
incorrect segments during training. 
We leave this for future work.

\section*{Acknowledgements}

We would like to thank the anonymous reviewers for their valuable comments. Wei Lu is supported by Singapore Ministry of Education Academic Research Fund (AcRF) Tier 2 Project MOE2017-T2-1-156, and is partially supported by SUTD project PIE-SGP-AI-2018-01.

\bibliography{disco}
\bibliographystyle{acl_natbib}

\appendix

\section{Segment Extraction}

Neural segmental hypergraphs \cite{D18-1019} were proposed for modeling overlapping structures in entity mentions. 
We directly adopt their approach to model the segments of overlapping structures. 
Note that our segment also holds the information of entity type, 
so the resulting system for segment extraction can also be viewed as performing sub-mention recognition. 
Next, we illustrate how the segmental hypergraph encodes the overlapping segments by a concrete example. 
For brevity, we only show the example that is annotated with one entity type, and it is able to be trivially extended to the case of multiple entity types. 

Given a phrase ``He had blood in his mouth and on his tongue", there exist two disorder mentions: 
`\textit{blood in his mouth}' and `\textit{blood ... on his tongue}' where the second mention has a discontiguous sequence of words.
Our two-stage approach first extracts segments that lead to these two mentions.
In this example, the segments consist of `\textit{blood}', `\textit{blood in his mouth}' and `\textit{on his tongue}'. 
We observe that the first two segments overlap with each other.

Segmental hypergraph encodes this segment combination based on five types of nodes:

\begin{itemize}
\item $\boldsymbol{\mathsf{A}}_i$ encodes all segments that start with the $i$-th or a later word
\item $\boldsymbol{\mathsf{E}}_i$ encodes all  segments that start exactly with the $i$-th word
\item $\boldsymbol{\mathsf{T}}^k_i$ represents all segments of type $k$ starting with the $i$-{th} word
\item $\boldsymbol{\mathsf{I}}^k_{i,j}$ represents all segments of type $k$ that contain the $j$-th word and start with the $i$-th word
\item $\boldsymbol{\mathsf{X}}$ marks the end of a segment.
\end{itemize}

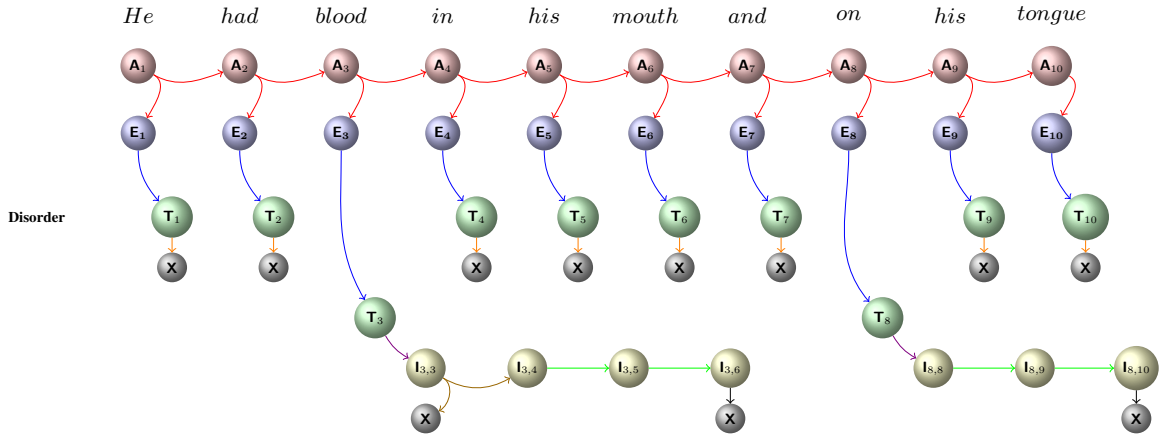
\begin{figure*}[t]
\small
\begin{center}
\begin{tikzpicture}[scale=0.5, every node/.style={scale=.70}]

\tikzset{VertexStyle/.style={
  shape=circle,
  shading=ball,
  ball color=red!20,
  minimum size=18pt}
}

\Vertex[x=0.0,y=6.0,L=$\boldsymbol{\mathsf{A}}_1$]{A1}
\Vertex[x=3.0,y=6.0,L=$\boldsymbol{\mathsf{A}}_{2}$]{A2}
\Vertex[x=6.0,y=6.0,L=$\boldsymbol{\mathsf{A}}_{3}$]{A3}
\Vertex[x=9.0,y=6.0,L=$\boldsymbol{\mathsf{A}}_{4}$]{A4}
\Vertex[x=12.0,y=6.0,L=$\boldsymbol{\mathsf{A}}_{5}$]{A5}
\Vertex[x=15.0,y=6.0,L=$\boldsymbol{\mathsf{A}}_{6}$]{A6}
\Vertex[x=18.0,y=6.0,L=$\boldsymbol{\mathsf{A}}_{7}$]{A7}
\Vertex[x=21.0,y=6.0,L=$\boldsymbol{\mathsf{A}}_{8}$]{A8}
\Vertex[x=24.0,y=6.0,L=$\boldsymbol{\mathsf{A}}_{9}$]{A9}
\Vertex[x=27.0,y=6.0,L=$\boldsymbol{\mathsf{A}}_{10}$]{A10}

\tikzset{VertexStyle/.style={
  shape=circle,
  shading=ball,
  ball color=blue!20,
  minimum size=18pt}
}

\Vertex[x=0.0,y=4.0,L=$\boldsymbol{\mathsf{E}_1}$]{E1}
\Vertex[x=3.0,y=4.0,L=$\boldsymbol{\mathsf{E}_{2}}$]{E2}
\Vertex[x=6.0,y=4.0,L=$\boldsymbol{\mathsf{E}_{3}}$]{E3}
\Vertex[x=9.0,y=4.0,L=$\boldsymbol{\mathsf{E}_{4}}$]{E4}
\Vertex[x=12.0,y=4.0,L=$\boldsymbol{\mathsf{E}_{5}}$]{E5}
\Vertex[x=15.0,y=4.0,L=$\boldsymbol{\mathsf{E}_{6}}$]{E6}
\Vertex[x=18.0,y=4.0,L=$\boldsymbol{\mathsf{E}_{7}}$]{E7}
\Vertex[x=21.0,y=4.0,L=$\boldsymbol{\mathsf{E}_{8}}$]{E8}
\Vertex[x=24.0,y=4.0,L=$\boldsymbol{\mathsf{E}_{9}}$]{E9}
\Vertex[x=27.0,y=4.0,L=$\boldsymbol{\mathsf{E}_{10}}$]{E10}

\tikzset{VertexStyle/.style={
  shape=circle,
  shading=ball,
  ball color=green!20,
  minimum size=18pt}
}

\Vertex[x=1.0,y=1.5,L=$\ \boldsymbol{\mathsf{T}}_1\ $]{T1b}
\Vertex[x=4.0,y=1.5,L=$\ \boldsymbol{\mathsf{T}}_2\ $]{T2b}
\Vertex[x=7.0,y=-1.5,L=$\ \boldsymbol{\mathsf{T}}_3\ $]{T3b}
\Vertex[x=10.0,y=1.5,L=$\ \boldsymbol{\mathsf{T}}_4\ $]{T4b}
\Vertex[x=13.0,y=1.5,L=$\ \boldsymbol{\mathsf{T}}_5\ $]{T5b}
\Vertex[x=16.0,y=1.5,L=$\ \boldsymbol{\mathsf{T}}_6\ $]{T6b}
\Vertex[x=19.0,y=1.5,L=$\ \boldsymbol{\mathsf{T}}_7\ $]{T7b}
\Vertex[x=22.0,y=-1.5,L=$\ \boldsymbol{\mathsf{T}}_8\ $]{T8b}
\Vertex[x=25.0,y=1.5,L=$\ \boldsymbol{\mathsf{T}}_9\ $]{T9b}
\Vertex[x=28.0,y=1.5,L=$\ \boldsymbol{\mathsf{T}}_{10}\ $]{T10b}

\tikzset{VertexStyle/.style={
  shape=circle,
  shading=ball,
  ball color=gray!50,
  minimum size=18pt}
}

\Vertex[x=1.0,y=-0,L=$\boldsymbol{\mathsf{X}}$]{1X}
\Vertex[x=4.0,y=-0,L=$\boldsymbol{\mathsf{X}}$]{2X}
\Vertex[x=10.0,y=-0,L=$\boldsymbol{\mathsf{X}}$]{4X}
\Vertex[x=13.0,y=-0,L=$\boldsymbol{\mathsf{X}}$]{5X}
\Vertex[x=16.0,y=-0,L=$\boldsymbol{\mathsf{X}}$]{6X}
\Vertex[x=19.0,y=-0,L=$\boldsymbol{\mathsf{X}}$]{7X}
\Vertex[x=25.0,y=-0,L=$\boldsymbol{\mathsf{X}}$]{9X}
\Vertex[x=28.0,y=-0,L=$\boldsymbol{\mathsf{X}}$]{10X}



\Vertex[x=8.5,y=-4.5,L=$\boldsymbol{\mathsf{X}}$]{X33}
\Vertex[x=17.5,y=-4.5,L=$\boldsymbol{\mathsf{X}}$]{X36}

\Vertex[x=29.5,y=-4.5,L=$\boldsymbol{\mathsf{X}}$]{X810}

\tikzset{VertexStyle/.style={
  shape=circle,
  shading=ball,
  ball color=yellow!20,
  minimum size=18pt}
}


\Vertex[x=8.5,y=-3,L=$\boldsymbol{\mathsf{I}}_{3,3}$]{I33b}
\Vertex[x=11.5,y=-3,L=$\boldsymbol{\mathsf{I}}_{3,4}$]{I34b}
\Vertex[x=14.5,y=-3,L=$\boldsymbol{\mathsf{I}}_{3,5}$]{I35b}
\Vertex[x=17.5,y=-3,L=$\boldsymbol{\mathsf{I}}_{3,6}$]{I36b}

\Vertex[x=23.5,y=-3,L=$\boldsymbol{\mathsf{I}}_{8,8}$]{I88b}
\Vertex[x=26.5,y=-3,L=$\boldsymbol{\mathsf{I}}_{8,9}$]{I89b}
\Vertex[x=29.5,y=-3,L=$\boldsymbol{\mathsf{I}}_{8,10}$]{I810b}

\tikzset{VertexStyle/.style = {line width=1pt,scale=1.5}}
\Vertex[x=0.0,y=7.5,L=$He$]{Time1}
\Vertex[x=3.0,y=7.5,L=$had$]{Time1}
\Vertex[x=6.0,y=7.5,L=$blood$]{Time2}
\Vertex[x=9.0,y=7.5,L=${in}$]{Time2}
\Vertex[x=12.0,y=7.5,L=$his$]{Time3}
\Vertex[x=15.0,y=7.5,L=$mouth$]{Time4}
\Vertex[x=18.0,y=7.5,L=$and$]{Time4}
\Vertex[x=21.0,y=7.5,L=$on$]{Time4}
\Vertex[x=24.0,y=7.5,L=$his$]{Time4}
\Vertex[x=27.0,y=7.5,L=$tongue$]{Time4}


\tikzset{VertexStyle/.style = {line width=1pt,scale=1,font=\bfseries}}
\Vertex[x=-3.0,y=1.5,L=$\textbf{Disorder}$]{LB1}


\draw[red][->]    (A1) to[out=-30,in=-160, draw=blue]  (A2);
\draw[red][->]    (A1) to[out=-30,in=60] (E1);

\draw[red][->]    (A2) to[out=-30,in=-160, draw=blue]  (A3);
\draw[red][->]    (A2) to[out=-30,in=60] (E2);

\draw[red][->]    (A3) to[out=-30,in=-160, draw=blue]  (A4);
\draw[red][->]    (A3) to[out=-30,in=60] (E3);

\draw[red][->]    (A4) to[out=-30,in=-160, draw=blue]  (A5);
\draw[red][->]    (A4) to[out=-30,in=60] (E4);

\draw[red][->]    (A5) to[out=-30,in=-160, draw=blue]  (A6);
\draw[red][->]    (A5) to[out=-30,in=60] (E5);

\draw[red][->]    (A6) to[out=-30,in=-160, draw=blue]  (A7);
\draw[red][->]    (A6) to[out=-30,in=60] (E6);

\draw[red][->]    (A7) to[out=-30,in=-160, draw=blue]  (A8);
\draw[red][->]    (A7) to[out=-30,in=60] (E7);

\draw[red][->]    (A8) to[out=-30,in=-160, draw=blue]  (A9);
\draw[red][->]    (A8) to[out=-30,in=60] (E8);
\draw[red][->]    (A9) to[out=-30,in=60] (E9);

\draw[red][->]    (A9) to[out=-30,in=-160, draw=blue]  (A10);
\draw[red][->]    (A10) to[out=-30,in=60] (E10);

\draw[blue][->]    (E1) to[out=-90,in=130] (T1b);
\draw[blue][->]    (E2) to[out=-90,in=130] (T2b);
\draw[blue][->]    (E3) to[out=-90,in=120] (T3b);
\draw[blue][->]    (E4) to[out=-90,in=130] (T4b);
\draw[blue][->]    (E5) to[out=-90,in=130] (T5b);
\draw[blue][->]    (E6) to[out=-90,in=130] (T6b);
\draw[blue][->]    (E7) to[out=-90,in=130] (T7b);
\draw[blue][->]    (E8) to[out=-90,in=130] (T8b);
\draw[blue][->]    (E9) to[out=-90,in=130] (T9b);
\draw[blue][->]    (E10) to[out=-90,in=130] (T10b);

\draw[orange][->]     (T1b) to[out=-90,in=90] (1X);
\draw[orange][->]     (T2b) to[out=-90,in=90] (2X);
\draw[orange][->]     (T4b) to[out=-90,in=90] (4X);
\draw[orange][->]     (T5b) to[out=-90,in=90] (5X);
\draw[orange][->]     (T6b) to[out=-90,in=90] (6X);
\draw[orange][->]     (T7b) to[out=-90,in=90] (7X);
\draw[orange][->]     (T9b) to[out=-90,in=90] (9X);
\draw[orange][->]     (T10b) to[out=-90,in=90] (10X);

\draw[blue!50!red][->]     (T3b) to[out=-60,in=150] (I33b);
\draw[blue!50!red][->]     (T8b) to[out=-60,in=150] (I88b);

\draw[green][->]    (I34b) to[out=0,in=180] (I35b);
\draw[green][->]    (I35b) to[out=0,in=180] (I36b);

\draw[green][->]    (I88b) to[out=0,in=180] (I89b);
\draw[green][->]    (I89b) to[out=0,in=180] (I810b);

%
\draw[green!40!red][->]    (I33b) to[out=-30,in=-150] (I34b);
\draw[black][->]    (I36b) to[out=-90,in=90] (X36);
\draw[black][->]    (I810b) to[out=-90,in=90] (X810);


\draw[green!40!red][->]   (I33b) to[out=-30,in=30] (X33);

\end{tikzpicture}
\caption{
A hyperpath that encodes three mention segments: `\textit{blood}', `\textit{blood in his mouth}' and `\textit{on his tongue}'.
}
\label{fig:hg}
\end{center}
\end{figure*}

Each segment can be expressed in terms of these five nodes and corresponds with a path in the segmental hypergraph. 
As a result, each segment combination corresponds with a \textit{hyperpath} where hyperedges are designated to connect multiple nodes so as to model overlapping segments.
Figure \ref{fig:hg} shows such a hyperpath for the segment combination in our example phrase.
Since we only have one entity type in this example, 
we eliminate the superscript $k$ in $\boldsymbol{\mathsf{T}}$ and $\boldsymbol{\mathsf{I}}$ nodes that indicates the information of entity type.

Starting from the third word `blood', there exist two segments `\textit{blood}' and `\textit{blood in his mouth}'.
The brown hyperedge with the parent node being  $\boldsymbol{\mathsf{I}}_{3,3}$ is responsible for connecting these two overlapping segments.
This hyperedge means that there exists a segment that ends at the third word (the link from $\boldsymbol{\mathsf{I}}_{3,3}$  to $\boldsymbol{\mathsf{X}}$) 
and there also exists a segment that continues to the next word (the link from $\boldsymbol{\mathsf{I}}_{3,3}$  to $\boldsymbol{\mathsf{I}}_{3,4}$ ).
The segment `\textit{on his tongue}' is directly mapped to the path from $\boldsymbol{\mathsf{T}}_8$ to  $\boldsymbol{\mathsf{X}}$.

The score for each hyperpath is the sum of the scores that are computed over each hyperedge. 
Since $\boldsymbol{\mathsf{T}}$ nodes encode word-level information and $\boldsymbol{\mathsf{I}}$ nodes encode span-level information,
two LSTMs are employed to capture the interactions at both word level and span level respectively. 
We use their original implementation that is publicly available\footnote{\url{https://github.com/berlino/overlapping-ner-em18}}.
\\
\\


\section{Heuristics for Handling Ambiguity }

\begin{figure}
\includegraphics[width=7.7cm]{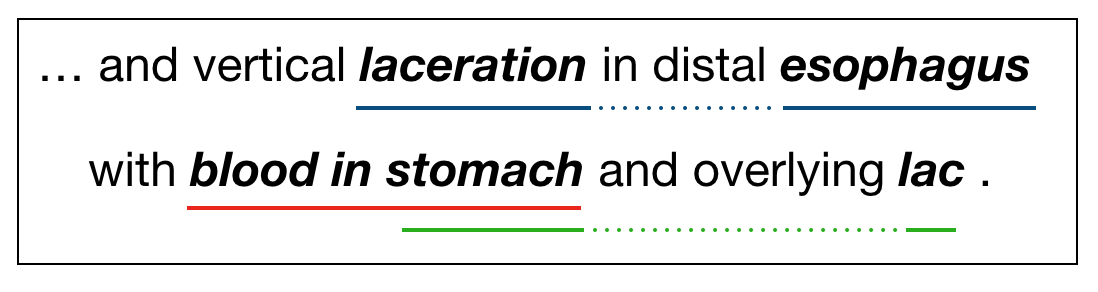}
\caption{Entities are highlighted with colored underlines. ``{\em laceration ... esophagus}" and ``{\em stomach ... lac}" contain discontiguous sequence of words and the latter also overlaps with another entity ``{\em blood in stomach}". }
\label{fig:example}
\end{figure}

\begin{figure}
\includegraphics[width=7.7cm]{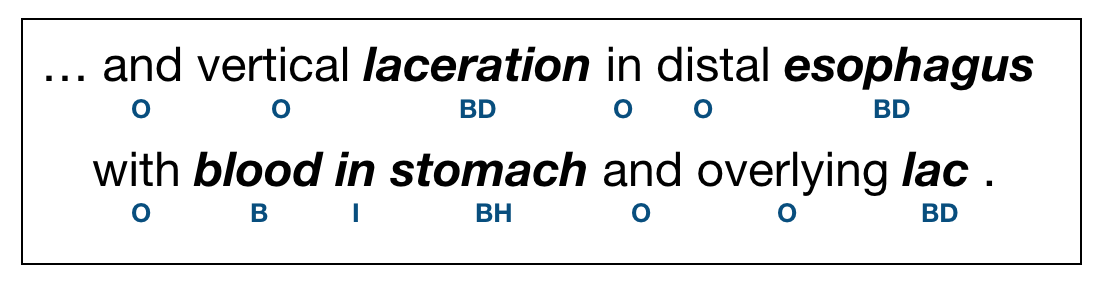}
\caption{Entities annotated using seven tags. }
\label{fig:tag}
\end{figure}

This section tries to explain the two heuristics ``enough" and ``all" when ambiguous tag sequences occur. 
We use the extended BIO tagging scheme  \cite{tang2013recognizing,muis2016learning} for example.

To encode the three discontiguous entities in Figure \ref{fig:example}, this tagset has seven tags: 

\begin{itemize}
  \setlength\itemsep{0em}
\item B/I:  \textbf{B}eginning and  \textbf{I}nside of contiguous entities
\item BH/IH: \textbf{B}eginning and  \textbf{I}nside of  \textit{head}  where head refers to segments shared by multiple discontiguous entities.
\item BD/ID:  \textbf{B}eginning and  \textbf{I}nside of  \textit{body}  where body  refers to segments that are not shared across entities.
\item O:  \textbf{O}utside of entities.
\end{itemize}

The resulting tag sequence is shown in Figure \ref{fig:tag}. 
Since this tagging scheme cannot model the correspondence between different tags, 
tagging sequences are very likely to have multiple interpretations.
For instance, it is not clear that ``laceration'' should be combined with ``esophagus'' or with ``stomach''.

The ``all" heuristics extracts all the possible entities that could exist in the tagging sequence.
In this case, ``all" heuristics will produce ``{\em laceration ... esophagus}" , ``{\em stomach ... lac}", ``{\em blood in stomach}",  ``{\em laceration ... lac}", ``{\em esophagu ... lac}",
``{\em laceration ...esophagus ... lac}",  ``{\em laceration ... stomach}", ``{\em esophagus ... stomach}", ``{\em laceration... stomach ... lac}", ``{\em esophagus ... stomach...lac}".

The ``enough" heuristics tries to find the minimal set of entities that corresponds to this tagging sequence.
In this case, ``enough" heuristics would produce at least three entities like: ``{\em laceration ... esophagus}" , ``{\em stomach ... lac}" and ``{\em blood in stomach}"; 
``{\em laceration ... lac}" , ``{\em blood in stomach}" and ``{\em esophagus ... stomach}".
We make further constraints to generate only one combination following \citet{muissupplementary}.

\section{Hyperparameters}

The hyperparameters used in our neural two-stage model are listed in Table \ref{tab:hyper}.
Since the size of our dataset is relatively small, the dropout is crucial to prevent overfitting considering that the pre-traind word embeddings have the dimension of 200.
The length of most segments is not greater than 6, so we set the maximal length $c$ to be 6 to improve the efficiency of segment extraction.

We also tried to incorporate a character-level component \cite{lample2016neural} to capture morphological and orthographic information.
However, it does not have a significant effect on the performance in term of $F_1$.

\begin{table}[h]
  \centering
  \scalebox{0.9}
{
\begin{tabular}{l|c}
\hline
word embedding dim & 200 \\
LSTM(word) hidden size & 128  \\
LSTM(span) hidden size & 128  \\
LSTM(entity) hidden size & 64  \\
maximal length $c$ & 6 \\
dropout  & 0.8  \\
$l_2$ & 0.0001
\end{tabular}
}
\caption{Hyperparameters of our joint model.}
\label{tab:hyper}
\end{table}

\end{document}